\ifdefined\pdfminorversion\pdfminorversion=7\fi
\documentclass[letterpaper,10pt,conference]{ieeeconf}
\IEEEoverridecommandlockouts
\usepackage[T1]{fontenc}
\usepackage{graphicx}
\usepackage{amsmath,amssymb}
\usepackage{cite}
\usepackage{microtype}
\usepackage{booktabs}
\usepackage{url}
\usepackage{flushend}
\usepackage[hidelinks]{hyperref}
\hypersetup{
  pdftitle={Diffeomorphic Hidden-State Steering: Improving Imitation Learning Policy Robustness through Expert Hidden-State Geometry},
  pdfauthor={KwangBin Lee},
  pdfsubject={Test-time hidden-state steering for frozen imitation learning policies},
  pdfkeywords={robotics, imitation learning, hidden-state steering, robustness}
}

\newcommand{\R}{\mathbb{R}}
\newcommand{\E}{\mathbb{E}}
\newcommand{\norm}[1]{\left\lVert #1\right\rVert}

\newcommand{\figref}[1]{Fig.~\ref{#1}}
\newcommand{\eqnref}[1]{Eq.~\eqref{#1}}
\newcommand{\tabref}[1]{Table~\ref{#1}}
\newcommand{\secref}[1]{Section~\ref{#1}}
\newcommand{\appref}[1]{Appendix~\ref{#1}}

\newcounter{algorithm}
\newcommand{\algref}[1]{Algorithm~\ref{#1}}

\DeclareMathOperator{\diag}{diag}
\DeclareMathOperator*{\argmin}{arg\,min}

\title{\LARGE\bf Diffeomorphic Hidden-State Steering: Improving Imitation Learning Policy Robustness through Expert Hidden-State Geometry}
\author{KwangBin Lee\thanks{Correspondence: \texttt{lkb20015@kaist.ac.kr}.}}

\begin{document}
\maketitle

\thispagestyle{empty}
\pagestyle{empty}

\begin{abstract}
Small observation or action errors can compound during closed-loop execution and disrupt a trained imitation learning policy.
We study whether the original expert demonstrations can improve robustness through test-time hidden-state steering without additional demonstrations or policy retraining.
Our analysis of a trained policy finds that expert hidden-state variation is concentrated in relatively few dimensions, while the directions capturing this variation differ across regions.
Motivated by this geometry, we propose Diffeomorphic Hidden-State Steering (DHS), which learns an invertible nonlinear coordinate map from expert hidden states obtained by replaying the demonstrations through the frozen policy.
DHS attenuates coordinates with low expert variance more strongly than those with high variance, then maps the modified coordinates back to hidden-state space.
This allows steering directions to vary with the current hidden state while retaining every coordinate.
Across four simulated manipulation tasks, DHS improves mean success under observation noise over the unmodified Transformer policies.
Controlled comparisons yield higher mean success with learned nonlinear coordinates than with linear coordinates under both hard projection and full-rank rescaling.
Combining DHS with action smoothing improves mean success rate under observation noise in certain tasks compared with action smoothing alone.
\end{abstract}

\section{Introduction}
\label{sec:introduction}

Imitation learning (IL) trains robot policies from expert demonstrations, but the observations encountered during execution also depend on the policy's own actions.
Observation noise or action errors can therefore take the policy into conditions poorly represented in the demonstrations, where further errors can compound and disrupt task execution.
DAgger addresses this distribution mismatch by obtaining expert actions on learner-visited states, while DART collects demonstrations that include recovery from disturbances~\cite{ross2011dagger,laskey2017dart}.
Online reinforcement learning can also improve pretrained robot policies by updating their parameters using reward feedback from interaction~\cite{yang2023robot}.
These approaches all require information beyond the original demonstration set, such as new expert labels, additional demonstrations, or reward feedback from interaction.
This motivates a complementary question: whether the original demonstrations alone can improve the closed-loop robustness of a trained IL policy without additional demonstrations or policy retraining.

A neural IL policy processes observations into hidden states, which its later layers use to generate actions.
Replaying the original demonstrations through the frozen policy exposes the hidden states associated with expert behavior, providing a policy-specific reference that can be reused during deployment without additional supervision.
Intermediate hidden states therefore provide a natural intervention point: modifying one during execution can change the action while leaving the policy parameters fixed. We refer to this operation as \emph{test-time hidden-state steering}.
Prior work learns hidden-state correction modules jointly with the task network~\cite{lamb2019srn}, or steers trained robot policies using conceptors fitted to successful and failed roll-out activations~\cite{miao2026coast}.
In our demonstration-only setting, the steering reference instead consists solely of expert hidden states collected from the original demonstrations.
The resulting challenge is how to use the structure of these expert hidden states to suppress undesirable deviations while preserving the variation needed for action generation.

Our empirical analysis finds that expert hidden states vary predominantly within relatively few dimensions, while linear subspaces fitted to different clusters of expert hidden states differ in orientation (\secref{sec:geometry}).
These observations motivate viewing the expert distribution as concentrated near a low-dimensional manifold, with broad variation along local tangent subspaces and narrow variation along local normal subspaces.
Variation along this structure can reflect differences between demonstrated situations that require different actions.
The steering problem can therefore be viewed as reducing departures from the structure while limiting changes to the variation needed for action generation.

One way to pursue this goal is to attenuate components along directions with low expert variance more strongly than those along directions with high variance.
A linear filter can implement this rule using globally fitted directions and attenuation factors~\cite{jaeger2017conceptors}.
Under the manifold view, however, local tangent and normal subspaces can change orientation with location.
A fixed linear filter may therefore attenuate tangential expert variation as well as normal deviations, motivating steering directions that adapt to the local geometry.

\begin{figure*}[t]
\centering
\includegraphics[width=\textwidth]{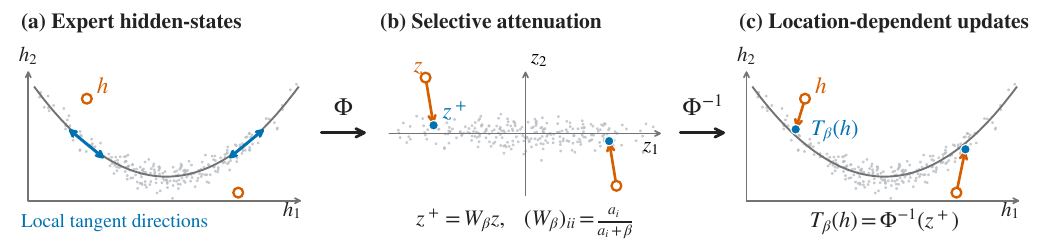}
\caption{Schematic overview of DHS.
(a) Expert hidden states are schematically shown near a curved structure; blue arrows indicate local tangent directions.
(b) A learned diffeomorphic chart maps hidden states to coordinates where DHS attenuates low-variance coordinates more strongly than high-variance coordinates while retaining every coordinate.
(c) Mapping back through the inverse chart produces location-dependent hidden-state updates.
Gray points denote expert samples; open orange and filled blue markers indicate states before and after steering.
Here, $a_i$ is the fitted expert variance of chart coordinate $i$, and $\beta$ controls steering strength.}
\label{fig:overview}
\end{figure*}

Our key idea is to use nonlinear expert coordinates for selective attenuation, limiting changes to high-variance coordinates while attenuating low-variance coordinates more strongly.
We implement this idea with Diffeomorphic Hidden-State Steering (DHS), which learns a diffeomorphic coordinate map, or \emph{chart}, from expert hidden states.
DHS builds on score-based pullback Riemannian geometry (SPRG)~\cite{diepeveen2025sprg} to learn the chart and an anisotropic Gaussian model of expert variation in its coordinates.
Under the manifold view, high and low-variance coordinates serve as proxies for tangential variation and normal deviations, respectively.
The inverse chart provides location-dependent coordinate directions for selective attenuation in hidden-state space.
At test time, DHS maps the current hidden state to chart coordinates, rescales them according to their fitted expert variances, and maps the result back before the remaining policy layers generate an action (\figref{fig:overview}).

Our contributions are as follows:
\begin{itemize}
\item We characterize expert hidden-state geometry in a trained IL policy, observing concentrated variation and differences in the orientation of linear subspaces fitted to different clusters of expert hidden states.
\item We introduce DHS, a full-rank nonlinear steering operator that uses expert hidden-state geometry to attenuate low-variance chart coordinates while limiting changes to high-variance coordinates in a frozen IL policy.
\item We demonstrate that DHS achieves higher mean success than the unmodified policy under observation noise across four simulated manipulation tasks.
Also, learned nonlinear coordinates yield higher mean success than linear coordinates from principal component analysis (PCA) under both hard projection and full-rank rescaling.
\end{itemize}

\section{Related Work}
\label{sec:related}

\subsection{Hidden-State Steering and Filtering}
Hidden-state steering can be learned jointly with a task network or derived from the statistics of reference activations.
State-Reification Networks jointly train denoising or attractor modules with the task network to improve generalization, and combine state reification with adversarial training to improve adversarial robustness~\cite{lamb2019srn}.
The statistical approach underlies conceptor filters, which use second moments of recurrent activity to preserve characteristic patterns and control their generation~\cite{jaeger2017conceptors}.
COAST applies this principle to robot policies by combining conceptors fitted to successful and failed roll-out activations~\cite{miao2026coast}.
Each fitted conceptor applies a linear map in hidden-state space. DHS uses nonlinear expert coordinates whose induced coordinate basis can vary with the current hidden state, allowing variance-dependent steering to follow the learned geometry.

\subsection{Representation Geometry and Nonlinear Steering}
Studies of intrinsic dimension in image classification networks~\cite{ansuini2019intrinsic} and representation geometry in protein and image Transformers~\cite{valeriani2023geometry} motivate examining how expert variation is organized within robot policies.
A constructive approach to such geometry is provided in~\cite{diepeveen2025sprg}: anisotropic flows yield tractable pullback geometry and a Riemannian autoencoder that reconstructs data after retaining high-variance coordinates.
DHS builds on this coordinate construction for full-rank selective attenuation during policy execution.

Geometry also affects how interventions move between represented behaviors.
Interpolation along fitted activation manifolds is examined in Manifold Steering through behavioral transitions in structured language tasks and a visual world model~\cite{wurgaft2026manifold}.
For targeted behavioral changes, Curveball Steering applies contrastive shifts in polynomial kernel PCA coordinates, using approximate inversion and restoring the reconstruction residual~\cite{raval2026curveball}.
These approaches establish precedents for nonlinear geometric intervention; our focus is how expert variation can guide corrective attenuation steering while preserving execution of an existing robot task.

\subsection{Predictive Guidance and Activation Feedback}
Guidance can also be derived from an action's predicted consequences.
Proximity to expert observation embeddings supplies the objective in Latent Policy Barrier, where dynamics learned from expert data and policy roll-outs guide action denoising through predicted future embeddings~\cite{sun2025lpb}.
Predicted future state values provide another criterion: Cosmos Policy learns action, state, and value predictions by fine-tuning a video model, and uses these values to select action proposals in its planning mode~\cite{kim2026cosmospolicy}.
Feedback can instead operate across network depth, as in WA-LQR, which locally linearizes diffusion Transformer blocks to track contrastive robustness feature targets~\cite{hong2026walqr}.
For DHS, the correction reference is the fitted distribution of expert hidden states: coordinate variances determine the attenuation directly, without future state prediction or contrastive feature tracking.

\section{Problem Formulation}
\label{sec:problem}

Let $\mathcal D_E$ denote the expert demonstrations used to train an IL policy $\pi_\theta$.
At control step $t$, the policy receives an observation history $\mathcal O_t$ and generates an action using its original action selection rule.
We seek to improve its closed-loop task success under observation noise by modifying an intermediate hidden state during execution, with all policy parameters $\theta$ fixed.

We select a single intermediate layer $\ell$ and denote the hidden state to be modified at control step $t$ by $h_t\in\R^d$, where $d$ is its dimension.
A steering map $T:\R^d\to\R^d$ replaces this hidden state with
\begin{equation}
h_t^+=T(h_t).
\label{eq:policy-interface}
\end{equation}
At the selected layer, all other token representations remain unchanged.
The policy then resumes its original computation and generates an action using its original action selection rule.
We denote the policy with this intervention by $\pi_{\theta,T}$.

Before deployment, we replay observation histories from $\mathcal D_E$ through the frozen policy and collect hidden states at the same layer and token position.
These expert hidden states form the dataset $\mathcal H_E$, which provides the fitting reference for $T$.

The objective is for $\pi_{\theta,T}$ to achieve higher task success than $\pi_\theta$ under observation noise.

\section{Diffeomorphic Hidden-State Steering}
\label{sec:method}

\subsection{Empirical Geometry of Expert Hidden States}
\label{sec:geometry}
Studies of image classifiers and Transformers have reported evidence of low-dimensional manifold structure in hidden representations~\cite{ansuini2019intrinsic,valeriani2023geometry}.
Motivated by these findings, we examine whether the expert hidden-state distribution from an IL robot policy is concentrated in relatively few dimensions and whether regional linear subspaces differ in orientation.
We select the Square manipulation task from Robomimic~\cite{mandlekar2022robomimic} as a representative task and use a Transformer policy trained by behavior cloning (BC-Transformer).
We analyze the approximately 27,000 expert hidden states that form $\mathcal H_E$, obtained by replaying the original training demonstrations through the frozen policy.

To quantify variance concentration, we apply principal component analysis (PCA) to the mean-centered states.
We compute the cumulative explained variance as the sum of the leading covariance eigenvalues divided by the sum of all eigenvalues.
Using the $90\%$ explained variance criterion employed in prior work~\cite{ansuini2019intrinsic}, we find that 29 principal components suffice in the ambient 512 dimensional space.
Thus, most observed expert variation can be captured by a low-dimensional linear subspace.

For a curved manifold, local tangent subspaces can change orientation relative to the fixed coordinate axes of the hidden-state space.
To examine this possibility, we partition the same states into eight clusters using $k$-means and fit a rank-29 PCA subspace within each cluster, using these regional subspaces as empirical proxies for local tangent subspaces.
We compare every pair of PCA subspaces using principal angles, computed as the inverse cosines of the singular values of the matrix of inner products between their orthonormal basis vectors.
Each pair yields 29 principal angles; all are zero for identical subspaces, while larger angles indicate greater differences in orientation.
Averaging all 29 principal angles over all 28 cluster pairs gives $44.7^\circ$, indicating differences in the orientations of the fitted regional subspaces.
Together, the variance concentration and regional orientation differences support a low-dimensional manifold-like view of expert hidden-state geometry.
These observations therefore motivate the use of a nonlinear chart that allows coordinate rescaling to act along location-dependent directions in hidden-state space.

\subsection{Learning an Expert Coordinate Chart}
To construct such coordinates, we adopt SPRG's combination of an invertible nonlinear chart and a learned anisotropic Gaussian~\cite{diepeveen2025sprg}.
We fit a full-dimensional diffeomorphism $\Phi:\R^d\to\R^d$, a smooth bijection with a smooth inverse, to $\mathcal H_E$.
For a generic hidden state $h\in\R^d$, let $z=\Phi(h)$.
We model expert coordinates by $\mathcal N(0,A)$, where $A=\diag(a_1,\ldots,a_d)$ and each $a_i>0$ is the fitted \emph{expert variance} of coordinate $i$.
The inverse $\Phi^{-1}$ maps modified coordinates back to hidden-state space, where the corresponding update directions can vary with position.
The fitted variances \(a_i\) determine how strongly each coordinate is attenuated.

The chart combines fixed invertible preprocessing with a nonlinear flow composed of smooth invertible affine coupling blocks~\cite{dinh2017realnvp}.
We learn the flow and $A$ jointly with the preprocessing fixed.
Let $J_\Phi(h)$ denote the chart Jacobian.
By the change of variables formula, the negative log-likelihood (NLL) of the expert hidden states, up to a constant, is
\begin{align}
\mathcal L_{\mathrm{NLL}}
&=\E_h\!\left[\tfrac12\Phi(h)^\top A^{-1}\Phi(h)
+\tfrac12\log\det A\right.\nonumber\\
&\hspace{4.1em}\left.-\log|\det J_\Phi(h)|\right],
\label{eq:nll}
\end{align}
where $\E_h$ averages over expert hidden states in $\mathcal H_E$.
Likelihood alone leaves coordinate scales undetermined: doubling one chart coordinate and quadrupling its fitted variance leaves the hidden-state likelihood unchanged (\appref{app:scale}).
These scales also determine the attenuation factors introduced below, making regularization part of the steering design.
Following SPRG~\cite{diepeveen2025sprg}, we fit the chart with isometry and volume regularization:
\begin{equation}
\mathcal L_{\mathrm{fit}}
=\mathcal L_{\mathrm{NLL}}+\lambda_{\mathrm{iso}}\mathcal R_{\mathrm{iso}}
+\lambda_{\mathrm{vol}}\mathcal R_{\mathrm{vol}}.
\label{eq:fitting}
\end{equation}
Here, $\mathcal R_{\mathrm{iso}}$ encourages local preservation of Euclidean lengths by the trainable flow in preprocessed coordinates, while \(\mathcal R_{\mathrm{vol}}\) penalizes local volume expansion and contraction through its squared log absolute Jacobian determinant.
The regularization weights are nonnegative.
The preprocessing and penalty definitions are given in \appref{app:fitting}.

\subsection{Steering by Variance-Dependent Rescaling}

At control step $t$, let $z_t=\Phi(h_t)$.
With $A$ fixed after fitting, the negative log-likelihood of candidate chart coordinates \(r\) under the fitted Gaussian is \(\frac12 r^\top A^{-1}r\) up to a constant.
We balance this penalty against the change from $z_t$:
\begin{equation}
z_t^+=\argmin_{r\in\R^d}
\left\{\tfrac12\norm{r-z_t}_2^2+\tfrac\beta2 r^\top A^{-1}r\right\},
\qquad\beta\geq0.
\label{eq:proximal}
\end{equation}
The hyperparameter $\beta$ controls steering strength for a given perturbation condition and remains fixed within each roll-out.
The unique minimizer gives
\begin{align}
z_t^+&=W_\beta z_t,&W_\beta&=\diag\!\left(\frac{a_i}{a_i+\beta}\right)_{i=1}^d,\nonumber\\
T_\beta(h)&=\Phi^{-1}\!\left(W_\beta\Phi(h)\right).
\label{eq:dhs}
\end{align}
For steering strength $\beta>0$, coordinates with $a_i\gg\beta$ undergo small relative changes, while those with $a_i\ll\beta$ are strongly attenuated.
Every multiplier is strictly positive for finite $\beta$, so all coordinates are retained and $T_\beta$ is invertible in exact arithmetic.
At $\beta=0$, $T_0(h)=h$.
The strength $\beta$ therefore controls attenuation continuously without requiring a choice of retained dimension.

We replace $h_t$ with $h_t^+{\,=\,}T_\beta(h_t)$ at the intervention layer and resume the frozen policy as defined in \secref{sec:problem}.
\algref{alg:dhs} summarizes offline fitting and online execution.

\subsection{Location-Dependent Steering Directions}
Although $W_\beta$ is fixed, the inverse chart determines how coordinate changes are expressed locally in hidden-state space.
Let $e_i$ denote the $i$th standard basis vector in chart coordinates.
The corresponding coordinate basis vector at $h$ is
\begin{equation}
v_i(h)=J_\Phi(h)^{-1}e_i.
\label{eq:directions}
\end{equation}
For an affine chart, these basis vectors are constant, so steering acts along fixed hidden-state directions.
For a nonlinear chart, they can vary with $h$, allowing the same coordinate rescaling to act along directions that vary with position.
These basis vectors describe local changes; deployment evaluates the full nonlinear map in \eqnref{eq:dhs} with a forward chart evaluation, diagonal rescaling, and an inverse chart evaluation, without Jacobian evaluations or iterative optimization.
\appref{app:analysis} quantifies this selective attenuation and gives a conditional mean interpretation under an auxiliary Gaussian noise model.

\begin{center}
\begin{minipage}{0.97\columnwidth}
\small
\refstepcounter{algorithm}\label{alg:dhs}
\hrule\smallskip
\textbf{Algorithm \thealgorithm: DHS fitting and execution}\par
\smallskip
\noindent\textbf{Input:} $\mathcal D_E$, frozen $\pi_\theta$, intervention layer $\ell$, $\beta$, $\lambda_{\mathrm{iso}}$, $\lambda_{\mathrm{vol}}$.\par
\smallskip
\noindent
\begin{tabular}{@{}r@{\quad}p{0.90\linewidth}@{}}
\multicolumn{2}{@{}l}{\textbf{Offline fitting}}\\
1: & Replay $\mathcal D_E$ through $\pi_\theta$ to collect $\mathcal H_E$.\\
2: & Fit $\Phi,A$ to $\mathcal H_E$ using \eqnref{eq:fitting}; freeze $\Phi,A$.\\
3: & $W_\beta\leftarrow\diag\!\left(\frac{a_i}{a_i+\beta}\right)_{i=1}^d$.\\[2pt]
\multicolumn{2}{@{}l}{\textbf{Online execution}}\\
4: & \textbf{for} each control step $t$ \textbf{do}\\
5: & \quad Compute $h_t$ from $\mathcal O_t$ using $\pi_\theta$ through layer $\ell$.\\
6: & \quad $z_t\leftarrow\Phi(h_t)$; $z_t^+\leftarrow W_\beta z_t$.\\
7: & \quad $h_t^+\leftarrow\Phi^{-1}(z_t^+)$.\\
8: & \quad Replace only $h_t$ by $h_t^+$; resume $\pi_\theta$.\\
9: & \quad Execute the policy's action.\\
10: & \textbf{end for}
\end{tabular}
\par\smallskip\hrule
\end{minipage}
\end{center}

\section{Experiments}
\label{sec:experiments}

\begin{figure*}[t]
\centering
\includegraphics[width=\textwidth]{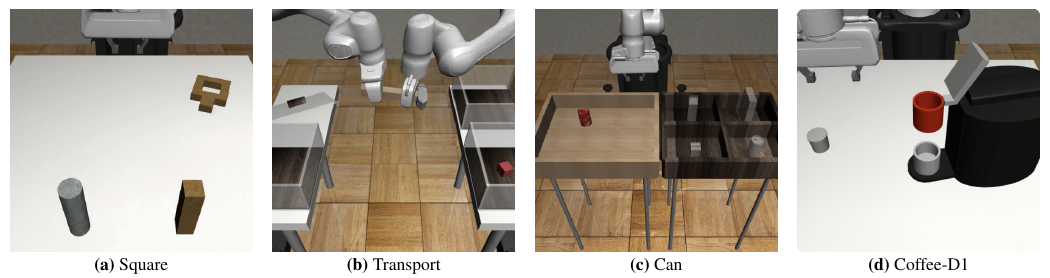}
\caption{Manipulation tasks used in evaluation: (a) Square, (b) Transport, (c) Can, and (d) Coffee-D1.}
\label{fig:tasks}
\end{figure*}

\begin{table*}[t]
\caption{Closed-loop task success ($\%$), mean $\pm$ SD across three roll-out seeds with 100 episodes per seed.}
\label{tab:main}
\centering
\footnotesize
\setlength{\tabcolsep}{3.4pt}
\begin{tabular}{lcccccccc}
\toprule
& \multicolumn{2}{c}{Square} & \multicolumn{2}{c}{Transport} & \multicolumn{2}{c}{Can} & \multicolumn{2}{c}{Coffee-D1}\\
\cmidrule(lr){2-3}\cmidrule(lr){4-5}\cmidrule(lr){6-7}\cmidrule(lr){8-9}
Method & Clean & $\sigma=0.05$ & Clean & $\sigma=0.02$ & Clean & $\sigma=0.05$ & Clean & $\sigma=0.01$\\
\midrule
No steering & $79.0\pm3.5$ & $39.7\pm3.1$ & $45.0\pm1.0$ & $37.3\pm2.1$ & $90.7\pm0.6$ & $55.3\pm6.7$ & $72.0\pm2.6$ & $67.0\pm7.9$\\
PCA projection & $66.7\pm3.8$ & $21.7\pm2.1$ & $24.3\pm2.9$ & $22.0\pm5.3$ & $82.0\pm4.6$ & $16.3\pm1.5$ & $17.7\pm2.1$ & $24.3\pm1.5$\\
AE reconstruction & $80.0\pm3.6$ & $32.7\pm4.5$ & $41.3\pm4.0$ & $43.3\pm7.5$ & $87.3\pm0.6$ & $54.0\pm2.6$ & $52.3\pm4.6$ & $53.0\pm3.5$\\
kNN replacement ($k=1$) & $53.7\pm7.4$ & $25.0\pm6.2$ & $13.7\pm4.6$ & $19.0\pm3.0$ & $71.0\pm3.0$ & $37.0\pm8.2$ & $32.3\pm1.2$ & $41.3\pm0.6$\\
kNN replacement ($k=5$) & $53.7\pm6.8$ & $23.7\pm6.1$ & $13.3\pm4.0$ & $17.7\pm1.5$ & $69.3\pm7.6$ & $34.7\pm5.7$ & $34.0\pm5.6$ & $36.7\pm3.8$\\
Gaussian score step & $\mathbf{82.3\pm1.5}$ & $42.3\pm6.0$ & $41.3\pm6.4$ & $38.0\pm4.6$ & $92.7\pm2.1$ & $57.7\pm3.5$ & $72.7\pm6.4$ & $69.7\pm3.1$\\
\textbf{DHS} & $78.7\pm0.6$ & $51.7\pm2.5$ & $36.0\pm3.0$ & $41.0\pm2.6$ & $\mathbf{93.7\pm0.6}$ & $\mathbf{61.7\pm1.5}$ & $\mathbf{77.3\pm4.9}$ & $\mathbf{73.0\pm3.5}$\\
\midrule
Action EWMA & $78.0\pm2.6$ & $57.7\pm7.0$ & $60.7\pm7.6$ & $56.3\pm3.8$ & $93.3\pm1.5$ & $57.3\pm0.6$ & $73.7\pm5.0$ & $61.3\pm2.5$\\
\textbf{DHS + Action EWMA} & $82.0\pm1.7$ & $\mathbf{61.7\pm3.1}$ & $\mathbf{63.3\pm2.3}$ & $\mathbf{62.0\pm1.7}$ & $92.0\pm3.0$ & $58.0\pm2.0$ & $69.7\pm4.0$ & $59.7\pm4.7$\\
\bottomrule
\end{tabular}
\end{table*}

\subsection{Evaluation Setup}
We evaluate policies on four simulated manipulation tasks: Square, Transport, and Can from Robomimic~\cite{mandlekar2022robomimic}, and Coffee-D1 from MimicGen~\cite{mandlekar2023mimicgen} (\figref{fig:tasks}).
For each task, all methods use the same frozen BC-Transformer checkpoint with low-dimensional observations.
All hidden-state methods intervene on the final token at the same layer of the frozen policy.
At each policy call, independent noise drawn from $\mathcal N(0,\sigma^2)$ is added to every element of the observation history in its original feature units; overlapping observations are re-perturbed on subsequent calls.
The protocol uses neither observation normalization nor quaternion renormalization.

We compare DHS with alternatives fitted to expert hidden states at the same layer: \emph{PCA projection}, \emph{autoencoder (AE) reconstruction}, \emph{$k$-nearest neighbor (kNN) replacement}, and a \emph{Gaussian score step}.
These alternatives test linear truncation, nonlinear reconstruction, replacement by the mean of the $k$ nearest expert hidden states, and density-based correction in standardized hidden-state coordinates.

At the action level, \emph{Action EWMA} (exponentially weighted moving average) combines recent action predictions with exponentially decaying weights, resetting between episodes.
We evaluate it alone and in combination with DHS. The unmodified policy is denoted as \emph{no steering}.
Details are in \appref{app:experimental}.

\subsection{Closed-Loop Success under Observation Noise}
\label{sec:task-success}
We evaluate DHS and all baselines under observation noise with level $\sigma$.
For each task, we choose a noise level that noticeably reduces the success rate of the no steering baseline relative to clean observations ($\sigma=0$).
The comparison in \tabref{tab:main} uses three roll-out seeds with 100 episodes per seed and reports percent success as mean $\pm$ sample standard deviation (SD) across seeds.
Bold entries indicate the highest mean in each column.
Policy and chart checkpoints remain fixed across seeds, so SD measures roll-out variability.
The task-specific DHS strengths $\beta$ are selected by hyperparameter search and used unchanged for clean and noisy evaluations, both with and without Action EWMA.

DHS improves mean noisy success over no steering on all four tasks in \tabref{tab:main}, by $3.7$ to $12.0$ percentage points.
It also achieves higher noisy means than the Gaussian score step on all four tasks and the highest noisy mean among the hidden-state methods on Square, Can, and Coffee-D1.
To examine the role of coordinate choice, \secref{sec:coordinate-controls} compares linear and nonlinear coordinates under hard projection and full-rank rescaling.

The interventions act at different stages: DHS adjusts the selected hidden-state before the current action is generated, while Action EWMA combines recent action predictions.
Adding DHS to Action EWMA increases mean noisy success from $57.7\%$ to $61.7\%$ on Square and from $56.3\%$ to $62.0\%$ on Transport, yielding the highest noisy mean among the evaluated methods on both tasks.
On Can and Coffee-D1, DHS alone achieves the highest noisy mean.
The gains over Action EWMA on Square and Transport show that hidden-state steering can also benefit policies whose action outputs are already smoothed.

\subsection{Why Use Learned Nonlinear Coordinates?}
\label{sec:coordinate-controls}
\secref{sec:geometry} shows that 29 principal components explain 90\% of the variance in expert hidden states on Square.
We now ask whether learned nonlinear coordinates are more effective than linear PCA coordinates for hidden-state steering.
We use Square as a representative task for this comparison under hard projection and full-rank rescaling, with PCA fitted to standardized expert hidden states.
Both comparisons use three roll-out seeds with 100 episodes per seed; noisy evaluations use a noise level of $\sigma{\,=\,}0.05$.

First, we use 29 as a common retained dimension for PCA and the learned chart.
For both PCA and the learned chart, we keep the 29 highest variance coordinates and set the remaining coordinates to zero.
We map the modified coordinates back to a hidden state and insert it at the same intervention layer of the frozen policy.
The learned chart yields higher mean success than PCA in both clean and noisy conditions (\tabref{tab:hard}), showing that the choice of coordinates affects task success even when the same number of coordinates is retained.

We next test whether the learned chart remains beneficial when all 512 coordinates are retained.
We compare DHS with \emph{PCA-Wiener}, which applies the same rescaling rule in the complete linear PCA basis using its empirical coordinate variances.
For this comparison, each method's strength is selected by mean development success and fixed before evaluation on disjoint seeds.
Both methods use matching episode initializations and random seeds.
DHS achieves higher mean noisy success, $50.3\pm9.1\%$ versus $44.0\pm4.6\%$, reported as mean $\pm$ sample SD across seeds (\tabref{tab:hard}).

The learned chart yields higher mean success under both update rules, supporting coordinate choice as an important component of effective hidden-state steering on Square.

\begin{table}[t]
\caption{Coordinate comparisons on Square: task success (\%), mean $\pm$ SD. Noisy evaluations use $\sigma=0.05$.}
\label{tab:hard}
\centering
\footnotesize
\renewcommand{\arraystretch}{1.12}
\begin{tabular*}{\columnwidth}
{@{}l@{\extracolsep{\fill}}cc@{}}
\toprule
\multicolumn{3}{@{}l}{
    \textit{Hard projection: 29 coordinates retained}
}\\[3pt]
Condition & PCA projection & Chart projection \\
\midrule
Clean & $66.7\pm3.8$ & $\mathbf{80.0\pm5.3}$ \\
Noisy & $21.7\pm2.1$ & $\mathbf{48.7\pm5.9}$ \\
\midrule
\multicolumn{3}{@{}l}{
    \textit{Full-rank rescaling: 512 coordinates retained}
}\\[3pt]
Condition
    & \shortstack{PCA-Wiener}
    & \shortstack{DHS} \\
\midrule
Noisy & $44.0\pm4.6$ & $\mathbf{50.3\pm9.1}$ \\
\bottomrule
\end{tabular*}
\par\vspace{3pt}

\end{table}

\subsection{Selective Attenuation and Steering Strength}
\label{sec:diagnostics}
\label{sec:rescaling}
We now quantify how strongly DHS attenuates low-variance chart coordinates and how much it changes high-variance chart coordinates during execution.
For $z{\,=\,}\Phi(h)$, let $z_{\mathrm{hi}}$ contain the $m$ coordinates with the largest fitted expert variances $a_i$ and $z_{\mathrm{lo}}$ contain the remaining coordinates.
We use $m=29$ on Square.
This split is used only for analysis; DHS retains all coordinates.
We define the \emph{low-variance chart norm} as the unweighted Euclidean norm of the low-variance block:
\begin{equation}
d_{\mathrm{lo}}(h)=\norm{z_{\mathrm{lo}}}_2.
\label{eq:low-variance-norm}
\end{equation}

We measure the update from $z$ to $z^+{\,=\,}W_\beta z$ over 13,492 intervention steps from noisy Square episodes at $\sigma{\,=\,}0.05$ and $\beta{\,=\,}0.3$.
The median low-variance chart norm decreases from $8.49$ before steering to $0.95$ after steering.
For the high-variance block, the median cosine similarity between $z_{\mathrm{hi}}$ and $z_{\mathrm{hi}}^+$ is $0.9989$.
Its median relative displacement, $\norm{z_{\mathrm{hi}}^+-z_{\mathrm{hi}}}_2/(\norm{z_{\mathrm{hi}}}_2+10^{-12})$, is $5.3\%$.
These measurements show that DHS strongly attenuates the low-variance block while making limited changes to the high-variance block during noisy Square roll-outs.

To relate selective attenuation to task success, we vary $\beta$ on Square at noise level $\sigma{\,=\,}0.05$, keeping the policy, chart, and fitted expert variances fixed.
Each strength uses three runs of 100 episodes, with percent success reported as mean $\pm$ sample SD across runs.
The low-variance chart norm after steering is reported as the mean of the run medians.

Across the five tested strengths from $\beta{\,=\,}0.1$ to $\beta{\,=\,}1.0$, mean success ranges from $48.0\%$ to $53.3\%$, above the $39.7\%$ no steering reference in \tabref{tab:main}.
Among the tested strengths, mean success is highest at $\beta{\,=\,}0.4$ and decreases for larger $\beta$, while the low-variance chart norm after steering continues to decrease (\tabref{tab:beta}).
Increasing $\beta$ strengthens attenuation in every chart coordinate, including high-variance coordinates (\eqnref{eq:dhs}).
This shows that stronger attenuation can impair execution despite further reducing the low-variance chart norm.

\begin{table}[t]
\caption{Sensitivity to steering strength on Square at $\sigma=0.05$.}
\label{tab:beta}
\centering
\footnotesize
\setlength{\tabcolsep}{7pt}
\begin{tabular}{rcr}
\toprule
$\beta$ & \shortstack{Success (\%)\\Mean $\pm$ SD} & \shortstack{$d_{\mathrm{lo}}$ after steering\\Mean of run medians}\\
\midrule
0.1 & $50.7\pm4.0$ & 2.323\\
0.2 & $52.3\pm3.5$ & 1.374\\
0.4 & $53.3\pm4.9$ & 0.766\\
0.5 & $49.3\pm3.8$ & 0.631\\
1.0 & $48.0\pm7.0$ & 0.334\\
\bottomrule
\end{tabular}
\end{table}

\subsection{Task Success under Variance Block Perturbations}
Following the attenuation measurements in \secref{sec:diagnostics}, we examine the consequences of perturbing high and low-variance chart coordinates on Square and Can.
We use the same variance blocks with $m{\,=\,}29$ and $m{\,=\,}11$, respectively.

At each injection step, two consecutive policy calls at the same environment state yield clean and noisy hidden states $h^{\mathrm c}$ and $h^{\mathrm n}$, with noise level $\sigma{\,=\,}0.05$.
Let $z^{\mathrm c}=\Phi(h^{\mathrm c})$ and $\delta z=\Phi(h^{\mathrm n})-z^{\mathrm c}$.
We add either variance block of $\delta z$ to $z^{\mathrm c}$ at its observed magnitude, leaving the other block clean.
We map back through $\Phi^{-1}$ and insert the hidden state at the same intervention layer.
Injection occurs at the first control step only or at every step, using three roll-out seeds with 100 episodes per seed for each condition.

With every-step injection, high-variance block perturbations yield lower mean success on both tasks; the difference between blocks is much smaller with first-step injection (\tabref{tab:components}).
The every-step comparison in PCA coordinates shows the same pattern: low-variance block injection yields higher mean success by $30.7$ percentage points on Square and $39.0$ on Can.
High expert variance therefore does not imply tolerance to observation noise.
DHS rescales the full noisy coordinate, so attenuation also changes its clean component.
For the paired clean and noisy coordinates defined above, $z^{\mathrm n}=z^{\mathrm c}+\delta z$ gives
\begin{equation}
W_\beta z^{\mathrm n}-z^{\mathrm c}
=(W_\beta-I)z^{\mathrm c}+W_\beta\delta z.
\label{eq:signal-perturbation}
\end{equation}
The first term is distortion of the clean coordinates; the second is the perturbation remaining after steering.
Limited attenuation of high-variance coordinates reduces their relative clean-component distortion, while leaving more of their perturbation.
The block-injection results measure task sensitivity to the observed perturbations; task-success comparisons evaluate the net effect of DHS.
Together with the strength sweep, these results highlight the tradeoff between perturbation attenuation and changes to coordinates used for action generation.

\begin{table}[t]
\caption{Task success (\%) under variance-block perturbations, mean $\pm$ SD across three roll-out seeds.}
\label{tab:components}
\centering
\footnotesize
\setlength{\tabcolsep}{3pt}
\begin{tabular}{llcc}
\toprule
Task & Injection & \shortstack{High-variance\\perturbation} & \shortstack{Low-variance\\perturbation}\\
\midrule
Square & First step & $80.3\pm2.5$ & $78.0\pm6.1$\\
Square & Every step & $47.7\pm1.5$ & $74.7\pm7.6$\\
Can & First step & $92.7\pm2.5$ & $93.0\pm3.5$\\
Can & Every step & $59.3\pm4.9$ & $89.3\pm2.9$\\
\bottomrule
\end{tabular}
\end{table}

\section{Discussion}
\label{sec:discussion}

\subsection{Expert Geometry as a Steering Reference}
DHS gives the original demonstrations a second role after policy training: their hidden states provide a reference for steering the frozen policy during execution.
This separates policy training from the design of the steering update, without requiring additional expert action targets.
The expert hidden-state density does not uniquely determine the update: different chart scales can describe the same density while yielding different DHS updates at a fixed $\beta>0$ (\appref{app:scale}).
Chart learning and regularization therefore help define the steering rule, in addition to fitting the expert distribution.

\subsection{Limitations}
DHS fits the distribution of expert hidden states, including their observation history information, without modeling temporal transitions.
A hidden-state can be compatible with the fitted distribution yet inappropriate for the current task phase or a successful continuation.
Combining expert geometry with predictions of an intervention's future consequences is a natural extension, connecting hidden-state steering with predictive guidance~\cite{sun2025lpb,kim2026cosmospolicy}.

The evaluation covers fixed BC-Transformer checkpoints with low-dimensional observations in simulation, and a single intervention site.
In \tabref{tab:main}, DHS reduces clean Transport success from $45.0\%$ to $36.0\%$, and adding Action EWMA to DHS lowers noisy success on Can and Coffee-D1.
The fitted geometry depends on preprocessing and chart regularization; its coordinate statistics provide a steering reference rather than a model of task error. Future work could also combine DHS with out-of-distribution (OOD) detection to adapt when and how strongly steering should be applied based on departures from the expert hidden-state distribution.

\section{Conclusion}
\label{sec:conclusion}

DHS uses expert hidden-state geometry to improve the robustness of frozen IL policies at test time.
Motivated by the low-dimensional manifold-like geometry of expert hidden states, its central idea is to apply variance-dependent rescaling in nonlinear coordinates to produce location-dependent steering directions.
We use SPRG-based chart learning to fit expert coordinates and their variances.
DHS improves mean success over the unmodified policy across four simulated manipulation tasks under observation noise.
On Square, the learned coordinate construction yields higher mean success than linear PCA coordinates under both hard projection and full-rank rescaling. The strength sweep further shows that stronger coordinate attenuation does not necessarily improve execution.
Together with the gains over unmodified policies and the benefits alongside Action EWMA on Square and Transport, these results establish expert hidden states as a useful reference for selective test-time steering without additional demonstrations or policy retraining.

\appendices
\section{Derivations for the Steering Operator}
\label{app:analysis}

\subsection{Hidden-State Displacements and Invertibility}
The chart determines how variance-dependent rescaling acts in hidden-state space.
For an affine chart $\Phi(h)=B(h-\mu)$ with invertible $B$, the steering map is $T_\beta(h)=\mu+B^{-1}W_\beta B(h-\mu)$, with fixed coordinate directions $B^{-1}e_i$.
For PCA-Wiener, let $\mu$ be the expert mean, $S$ a positive diagonal matrix of coordinate standard deviations, and $U\Lambda U^\top$ the covariance of standardized expert hidden states, with $U$ orthogonal and $\Lambda$ the diagonal matrix of PCA variances.
When all PCA variances are positive, setting $B=U^\top S^{-1}$ and $A=\Lambda$ gives
\begin{equation}
T_\beta^{\mathrm{PCA}}(h)
=\mu+SU\Lambda(\Lambda+\beta I)^{-1}U^\top S^{-1}(h-\mu).
\label{eq:pca-wiener}
\end{equation}
Its coordinate directions are the columns of $SU$, providing the fixed linear reference in \secref{sec:coordinate-controls}.

For a nonlinear chart, the basis vectors $v_i(h)$ in \eqnref{eq:directions} can vary with location.
Fix $h$ and $A\succ0$, and set $z=\Phi(h)$.
As $\beta\to0$, $W_\beta=I-\beta A^{-1}+O(\beta^2)$.
Taylor expansion of the smooth inverse chart, using $J_{\Phi^{-1}}(z)=J_\Phi(h)^{-1}$, yields
\begin{equation}
T_\beta(h)-h=-\beta\sum_{i=1}^d\frac{z_i}{a_i}v_i(h)+O(\beta^2).
\label{eq:field}
\end{equation}
This expresses the leading displacement in the coordinate basis at the starting state.

To extend this interpretation to finite strength, map the segment from $z$ to $W_\beta z$ through the inverse chart:
\begin{align}
\gamma_\beta(s)&=\Phi^{-1}\!\left([I+s(W_\beta-I)]z\right),\quad s\in[0,1],
\label{eq:finite-path}\\
\dot\gamma_\beta(s)&=J_\Phi(\gamma_\beta(s))^{-1}(W_\beta-I)z.
\label{eq:finite-path-tangent}
\end{align}
The second line follows by the chain rule.
Since $\gamma_\beta(0)=h$ and $\gamma_\beta(1)=T_\beta(h)$, integrating the tangent gives the exact displacement
\begin{equation}
T_\beta(h)-h
=-\sum_{i=1}^d\frac{\beta z_i}{a_i+\beta}\int_0^1v_i(\gamma_\beta(s))\,ds.
\label{eq:finite-displacement}
\end{equation}
The coefficients are fixed by the starting coordinates and steering strength; the basis vectors are evaluated along the path.
For an affine chart, the basis is constant along the path.
For a nonlinear chart, the basis vectors can change direction and magnitude; the integral captures this variation beyond the starting-state approximation in \eqnref{eq:field}.
Deployment evaluates the endpoint directly through \eqnref{eq:dhs}, without path integration or Jacobian evaluations.

This nonlinear change of coordinates preserves the full-rank property of $W_\beta$.
By the chain rule,
\begin{equation}
J_{T_\beta}(h)=J_\Phi(T_\beta(h))^{-1}W_\beta J_\Phi(h).
\label{eq:steering-jacobian}
\end{equation}
The chart Jacobians are invertible, and every diagonal entry of $W_\beta$ is positive for finite $\beta\geq0$, so $J_{T_\beta}(h)$ has rank $d$.
The explicit inverse $T_\beta^{-1}(h)=\Phi^{-1}(W_\beta^{-1}\Phi(h))$ establishes global invertibility.
Thus, DHS permits location-dependent steering while retaining every hidden-state dimension.

\subsection{Attenuation Bounds and Gaussian Conditional Mean}
\label{app:gaussian}
To quantify selective attenuation, order $a_1\geq\cdots\geq a_d>0$ and choose $1\leq m<d$.
Define $z_{\mathrm{hi}}=(z_1,\ldots,z_m)^\top$ and $z_{\mathrm{lo}}=(z_{m+1},\ldots,z_d)^\top$, with the same partition for $z^+=W_\beta z$.
At fixed steering strength $\beta\geq0$, the attenuation multiplier $a/(a+\beta)$ is non-decreasing in $a$, while the displacement multiplier $\beta/(a+\beta)$ is non-increasing.
The largest multipliers in the respective blocks therefore give
\begin{align}
\norm{z_{\mathrm{lo}}^+}_2
&\leq\frac{a_{m+1}}{a_{m+1}+\beta}\norm{z_{\mathrm{lo}}}_2,\nonumber\\
\norm{z_{\mathrm{hi}}^+-z_{\mathrm{hi}}}_2
&\leq\frac{\beta}{a_m+\beta}\norm{z_{\mathrm{hi}}}_2.
\label{eq:bounds}
\end{align}
Thus, $a_{m+1}\ll\beta\ll a_m$ permits strong low-variance attenuation and small relative high-variance displacement simultaneously.
Both bounds are tight, attained on coordinates $m+1$ and $m$, respectively.
They hold for every $z$; the partition is used only for analysis.

As $\beta$ increases, every attenuation multiplier $a_i/(a_i+\beta)$ decreases, including those in the high-variance block.
For each fixed $h$,
\begin{equation}
\lim_{\beta\to\infty}T_\beta(h)=\Phi^{-1}(0).
\label{eq:strong-steering-limit}
\end{equation}
Thus, large strengths suppress both blocks even though $T_\beta$ remains invertible for finite $\beta$.
The strength sweep in \secref{sec:rescaling} measures how attenuation and task success change at finite strengths.

The same weights arise in the auxiliary chart noise model $\widetilde z=z^\star+\epsilon$, with independent $z^\star\sim\mathcal N(0,A)$ and $\epsilon\sim\mathcal N(0,\sigma_z^2I)$, where $\sigma_z^2>0$.
The cross-covariance of $z^\star$ and $\widetilde z$ is $A$, and $\widetilde z$ has covariance $A+\sigma_z^2I$.
Gaussian conditioning gives
\begin{equation}
\E[z^\star\mid\widetilde z]
=A(A+\sigma_z^2I)^{-1}\widetilde z
=W_{\sigma_z^2}\widetilde z.
\label{eq:mmse}
\end{equation}
This conditional mean minimizes expected squared error in chart coordinates; smaller signal-to-noise ratios $a_i/\sigma_z^2$ yield stronger attenuation.
Here, $\sigma_z^2$ belongs to the chart noise model and is distinct from the observation noise variance $\sigma^2$.
For nonlinear $\Phi^{-1}$, mapping this conditional mean back need not give the conditional mean in hidden-state space.

\subsection{Scale Dependence}
\label{app:scale}
Let $D_c=\diag(c_1,\ldots,c_d)$ with $c_i>0$.
Replace $\Phi$ by $\Phi_c=D_c\Phi$ and $A$ by $A_c=D_cAD_c$.
If $p_A$ denotes the Gaussian density with covariance $A$, then
\begin{align}
p_{A_c}(D_cz)&=p_A(z)/\det D_c,\nonumber\\
|\det J_{\Phi_c}(h)|&=\det D_c\,|\det J_\Phi(h)|.
\label{eq:scale-density}
\end{align}
The two scale factors cancel in the change of variables density, so both parameterizations represent the same hidden-state distribution.
At fixed $\beta>0$, the rescaling factors nevertheless become $c_i^2a_i/(c_i^2a_i+\beta)$, which also determine the update after returning to the original chart coordinates.
Recovering the original map with a new strength $\beta'$ requires $\beta'=c_i^2\beta$ for every coordinate.
Thus, a common scale can be absorbed into the strength, whereas unequal coordinate scales cannot.
Equal likelihood therefore does not imply equivalent coordinate attenuation, even after selecting a scalar steering strength.
Preprocessing and regularization shape the relative coordinate scales on which the steering rule depends.

\section{Experimental Details}
\label{app:experimental}

\subsection{Policy and Chart Fitting}
\label{app:fitting}
Policies use six Transformer blocks and ten-observation contexts, trained with AdamW and learning rate linearly decayed from $10^{-4}$ to $10^{-5}$ over the first 100 epochs.
DHS modifies the 512-dimensional final token after zero-based layer 4.
For expert mean $\mu$, diagonal standard deviation matrix $S$, and full PCA basis $U$ of standardized expert hidden states,
\begin{align}
y&=U^\top S^{-1}(h-\mu),&\Phi(h)&=\psi(y),\nonumber\\
\Phi^{-1}(z)&=\mu+SU\psi^{-1}(z).
\label{eq:chart}
\end{align}
The trainable flow $\psi$ uses eight affine coupling blocks of hidden width 256, SiLU activations, log-scale clamp 2, and log-parameterized base variances.
Chart fitting uses AdamW at $10^{-4}$ for Square/Transport and $10^{-3}$ for Can/Coffee-D1; the latter use PCA warm-start, 300 epochs, and batch size 256.
Following SPRG~\cite{diepeveen2025sprg}, with $J=J_\psi(y)$,
\begin{align}
\mathcal R_{\mathrm{iso}}
&=\E_y\!\left[\norm{J^\top J-I}_F^2\right],
\label{eq:isometry}\\
\mathcal R_{\mathrm{vol}}
&=\E_y\!\left[(\log|\det J|)^2\right].
\label{eq:volume-regularizer}
\end{align}
We jointly fit $\psi,A$ using \eqnref{eq:fitting}, $\lambda_{\mathrm{iso}}=100$, $\lambda_{\mathrm{vol}}=2$, and Hutchinson estimation of $\mathcal R_{\mathrm{iso}}$.
Here, $\norm{\cdot}_F$ is the Frobenius norm; expectations use preprocessed expert states.

\subsection{Baseline Settings and Evaluation Protocol}
Hidden-state baselines operate in standardized coordinates; their outputs are unstandardized before the policy resumes.
The AE uses $512\!\to\!512\!\to\!d_{\mathrm{AE}}\!\to\!512\!\to\!512$, SiLU hidden activations, linear bottleneck/output layers, and mean squared reconstruction error.
PCA ranks and AE bottlenecks are 29, 32, 11, and 23 for Square, Transport, Can, and Coffee-D1, respectively.
kNN uses Euclidean distance; the Gaussian score step uses Ledoit--Wolf covariance and step scale $10^{-4}$.
Action EWMA averages up to five predictions with normalized weights proportional to $2^{-j}$ for lag $j$.

For \tabref{tab:main}, DHS strengths are $0.3$, $0.1$, $0.001$, and $0.001$ in that task order.
The paired Square comparison uses $\beta{\,=\,}0.4$ for DHS and $0.3$ for PCA-Wiener, selected by mean development success before evaluation on disjoint seeds.
A 20,000-resample paired bootstrap over 300 episode pairs gives a $95\%$ confidence interval of $[0.0,12.7]$ percentage points for DHS minus PCA-Wiener.
Attenuation measurements in \secref{sec:diagnostics} use Square with $\sigma{\,=\,}0.05$ and $\beta{\,=\,}0.3$.
The evaluations reported in \tabref{tab:main}, \tabref{tab:hard}, and \tabref{tab:beta} use different sets of roll-out episodes.

\section*{Acknowledgment}
This work originated from research conducted at KAIST.
The author thanks Prof.~SooJean Han and the Autonomous Control for Stochastic Systems (ACSS) Lab at KAIST for their support and the research environment in which this work began.
OpenAI's ChatGPT was used to assist with language polishing, and codes for implementation of the experiments.
The author takes full responsibility for all statements and final formulations.

\begingroup
\emergencystretch=1em
\bibliographystyle{ieeetr}
\bibliography{reference}
\endgroup
\end{document}